# Challenges and opportunities for machine learning in multiscale computational modeling

Phong C.H. Nguyen[1], Joseph B. Choi[1], H.S. Udaykumar[2,*], Stephen Baek[1,3,*]

[1]University of Virginia, School of Data Science, Charlottesville, VA 22904, United States

[2]University of Iowa, Department of Mechanical Engineering, Iowa City, IA 52242, United States

[3]University of Virginia, Department of Mechanical and Aerospace Engineering, Charlottesville, VA 22903, United States

*Corresponding Authors: hs-kumar@uiowa.edu, baek@virginia.edu

**Abstract:**

Many mechanical engineering applications call for multiscale computational modeling and simulation. However, solving for complex multiscale systems remains computationally onerous due to the high dimensionality of the solution space. Recently, machine learning (ML) has emerged as a promising solution that can either serve as a surrogate for, accelerate or augment traditional numerical methods. Pioneering work has demonstrated that ML provides solutions to governing systems of equations with comparable accuracy to those obtained using direct numerical methods, but with significantly faster computational speed. These high-speed, high-fidelity estimations can facilitate the solving of complex multiscale systems by providing a better initial solution to traditional solvers. This paper provides a perspective on the opportunities and challenges of using ML for complex multiscale modeling and simulation. We first outline the current state-of-the-art ML approaches for simulating multiscale systems and highlight some of the landmark developments. Next, we discuss current challenges for ML in multiscale computational modeling, such as the data and discretization dependence, interpretability, and data sharing and collaborative platform development. Finally, we suggest several potential research directions for the future.

## 1 Introduction

Multiscale computational modeling has emerged as a central part of many mechanical engineering applications in recent years. The need for multiscale models arises from the limitations of single-scale models in describing physical laws and phenomena at different spatial and/or temporal scales. Larger scale models alone cannot capture the fine-scale features or resolve key details of complex mechanical systems, while smaller scale models are computationally inefficient for simulating physical phenomena that manifest at a larger system scale. Multiscale modeling addresses this trade-off between physical fidelity and efficiency by using different numerical models to represent physical laws and phenomena at different scales and bridging scales using closure models. Over the past decade, a growing number of studies have adopted multiscale modeling and successfully demonstrated its ability to predict complex phenomena in areas such as fluid mechanics [1], [2], materials modeling and design [3], [4], manufacturing and process modeling [5], [6].

Broadly speaking, there are two main strategies for linking models at different scales. *Hierarchical* multiscale modeling approaches [5], [7]–[9], also known as *sequential* or *decoupled* multiscale modeling, precompute constitutive relations (e.g., elasticity constants, energy release rate) using fine-scale models. These fine-scale quantities of interest (QoIs) are typically computed 'offline' and then utilized during a system-scale computations that seek to model observable and measurable larger scale phenomena. On the contrary, *concurrent* multiscale modeling approaches compute the required QoIs for coarse-scale models

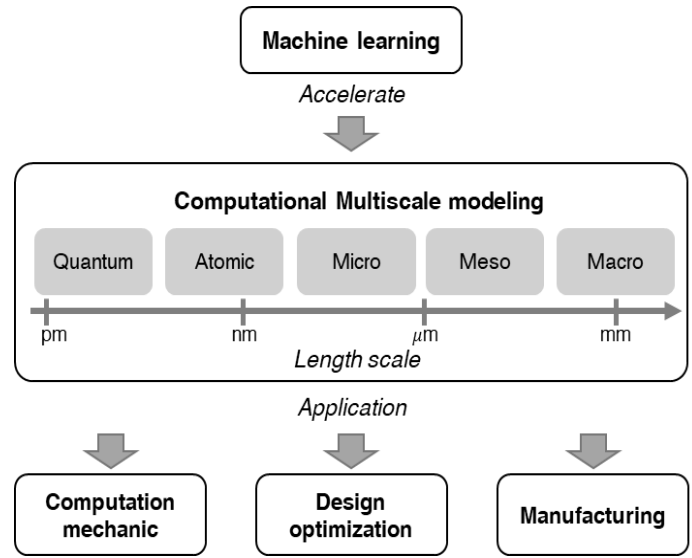

Figure 1. Machine learning can be applied to accelerate multiscale computational modeling. Examples of applications include computational mechanics, design optimization, and manufacturing modeling.

'on-the-fly,' as neeed, by coupling with the fine-scale models [10]–[13]. While hierarchical approaches require large amounts of memory and storage space to store precomputed QoIs, concurrent approaches face formidable computation costs as the problem size increases. These challenges limit the scalability and practicality of multiscale computational modeling in real-world applications.

Recent works suggest that machine learning (ML) has the potential to overcome the limitations of traditional multiscale modeling methods. For instance, to address the issue of memory and storage requirements in hierarchical modeling, ML techniques can be applied to learn a coarse, low fidelity, and low-cost representation, referred to as *representation learning*, of precomputed QoIs [14]. This allows for fast retrieval of information without the need for large memory and storage spaces. Concurrent multiscale modeling can also benefit from modern deep learning (DL) [15] models, which can replace traditional numerical solvers for 'on-the-fly' computation of QoIs. Previous studies have demonstrated that ML-based simulation can achieve comparable fidelity to direct numerical simulation while being considerably faster in computation time [16]. Moreover, ML can accelerate the solution of inverse problems, providing direct design solutions that meet design constraints without requiring the traditional iterative optimization process [17].

Despite the promise of using ML in multiscale modeling, several critical challenges have emerged from pioneering works. One of the most significant obstacles is the data dependency of ML, particularly with modern DL models. ML/DL has achieved remarkable success in computer vision by leveraging massive datasets with high-quality annotations, such as image labels and captions. However, this is not typically the case for multiscale modeling problems, which require intensive calculations and laborious data collection. Therefore, alongside the development of fast and efficient data generation methods, the research community may benefit from 'small data learning' techniques, such as transfer learning and meta learning (Sections 3.1.1 and 3.1.2), and collaborative/shared learning methods (Section 3.4).

Additionally, the majority of ML models are dependent on the discretization used during training, which can limit their scalability and practicality for multiscale problems where different discretization settings are used. This issue can be addressed by developing discretization-independent ML models such as those that employ continuous convolution or neural operators (Section 3.2), which are promising and warrant further investigation.

Lastly, the 'black-box' nature of ML models poses a challenge to their trustworthiness and interpretability in practical applications. While ML predictions can be accurate, the assimilated information during training that drive predictions may not always reflect the underlying physical phenomena. Hence, it is important to explain how predictions were derived and examine the physics-awareness of a model, in addition to the evaluation of their statistical performance. To achieve this, the development of techniques for interpreting ML predictions, such as neural network visualization (Section 3.3), is crucial and demands additional efforts.

This paper examines the promise of ML in accelerating the development of multiscale computational modeling and offers potential solutions to address the challenges. The paper is organized as follows. In Section 2, we present several examples of successful applications of ML in multiscale modeling. Section 3 discusses the challenges that previous works have revealed, as well as our perspective on potential solutions. Finally, we provide conclusions and remarks in Section 4.

## 2 Examples of machine learning in multiscale computational modeling

### 2.1 Computational mechanics

#### 2.1.1 Fluid modeling

The intersection of ML and fluid mechanics presents a natural synergy, owing to their similarity in dealing with large and complex datasets. In traditional fluid mechanics, one has to deal with a large amount of data coming from various source, including experiments, field measurements or large-scale simulation. The analysis of fluid mechanics data still relies largely on domain expertise and traditional statistical analysis, which involve many simplifying assumptions [18]. Recently, the rise of ML, especially DL, has been a significant driving force in fluid mechanics research, rendering opportunities for tackling more difficult fluid mechanics problems such as turbulent flows and its modeling. By leveraging the computational efficiency of ML, advancements in fluid flow modeling can be achieved, which can have a profound impact on a wide range of applications, such as biomedical engineering, aerospace engineering, and aerial/hydro-robotics and autonomous systems, among others.

In fact, DL techniques such as artificial neural networks (ANN) have been employed to simulate fluid dynamics very early, since the 1990s [19], [20] and are still being developed [21]. Along with the breakthrough in DL architecture development, the progress in fluid flow modeling has been significantly accelerated. ML approaches such as recurrent neural networks (RNNs) have been utilized to model flows in extreme environments [22], [23] Unlike traditional feedforward neural networks, RNNs have loops that allow information to be passed from one step of the network to the next, enabling them to maintain information about previous inputs and influence the current output. This characteristic makes RNNs capable of considering the inherent order of data and therefore well-suited to fluid mechanics modeling, where the data on the evolving flow fields is supplied to the machine learning algorithm in the form of sequences of images, videos, or time-series.

Another DL technique that has also demonstrated success in fluid modeling is Generative adversarial networks (GAN). The GAN framework is composed of two neural networks, namely the generator and the discriminator, that are trained in adversarial manner to generate high-quality synthetic data. Motivated by the capability of GAN to capture and generate realistic complex geometric patterns, Kim et al. [24] have applied GAN to model various complex fluid behaviors. These authors demonstrated that GAN is capable of accurately capturing the velocity field enabling many opportunities for further application including fast reconstruction of fluid flow, time-resampling, or fluid data compression.

Although current DL-based approaches have achieved successes, they have a limitation that trained models are typically only effective for interpolating data points and may not be as reliable when extrapolating beyond the known data range [18]. This reduces the generalizability and applicability of trained models and hinders their use in real-world applications. To address this limitation, it is crucial not only to expand the training dataset to ensure that most future prediction tasks fall within the interpolation of the training data but also to develop new modeling methods with better extrapolation capabilities.

### 2.1.2 Materials microstructure modeling

Materials microstructure modeling is another domain of application that has benefited from the advancement of ML. ML can greatly facilitate the materials microstructure modeling process by providing efficient and accurate methods for predicting materials properties and behavior of given material compositions [25] or microstructure [26]. ML can achieve this modeling capability due to its ability to identify and learn from data complex patterns that may be difficult to capture using traditional statistical methods. This modeling capability allows ML algorithms to learn complex relationships between a wide range of microstructure morphology parameters (e.g., grain size, grain orientation, or distribution of defects) and the output materials properties or mechanical response. These rapid and accurate predictions then can be used to design new materials with specific properties or to optimize the currently available materials with improved performance [27], [28].

Recent advances in DL, with the increase in approximation power of neural networks, has further accelerated progress in ML-based materials microstructure modeling. Modern DL methods such as convolutional neural networks (CNNs) can produce high-fidelity estimation of materials microstructure response to loads with an accuracy comparable to traditional numerical methods but with multiple order faster computation time [16], [29]. Physics-informed ML [30]–[32] (PIML) is another branch of DL method that has been widely adopted to simulate complex governing equations in multiscale materials modeling with increasing success. In PIML, prior physics knowledge is embedded in the design of DL algorithms, either via learning objectives [30], or architecture design [16], [31]. As reported in pioneering works [16], the embedded prior physics knowledge helps reduce the need for large datasets while maintaining the prediction accuracy and generalizability of the DL model. The strong predictive capabilities and computational efficiency of DL-based methods make them ideally suited as surrogates for design optimization [33], multiscale SPP linkage modeling [29], [34], [35], and uncertainty quantification tasks [36].

### 2.1.3 Reducing the nonlinearity of dynamical systems

Many complex mechanical systems are high nonlinear, posing a significant challenge for obtaining stable and reliable numerical solutions, which often requires an extremely fine grid resolution to resolve physics at features like shocks, interfaces and reactive fronts. The presence of nonlinearities and the potential for emergence of features at spatial and temporal hierarchies of scales renders the computation of such problems time-consuming. To address these challenges, the Koopman operator theory [37], [38], has been drawing attention recently; it is based on the idea that any highly nonlinear dynamical system can be represented in terms of an infinite-dimensional linear operator, called the Koopman operator. This linear representation of nonlinear dynamical systems facilitates the analysis, design, and control of complex system via well-established methods for linear system. Here, the challenge is in approximating the Koopman operator from data. This has led to improved numerical methods such as dynamic mode decomposition (DMD) [39], which utilized linear observables, and its extended version (eDMD) [40], which augments the original DMD with nonlinear observables. These methods have demonstrated successes in modeling nonlinear flows [39], the Duffing oscillator, and the Kuramoto Sivashinsky partial differential equation [40], among many others. DMD extracts rich and physically meaningful basis functions, which can notably improve the prediction accuracy of numerous classification and forecasting tasks involving time-series data [40], [41]. In addition to DMD and its variants, DL with the representation power of neural networks has been applied exclusively to model the Koopman-invariant subspaces of observables [38], [42], [43]. The advantage of the Koopman-based methods is that they do not require prior knowledge or an explicitly defined model, and therefore can be scalable to various applications with unknown physics.

## 2.2  Topology optimization

Topology optimization (TO) is another a group of computational design problems that benefit greatly from ML. TO determines the optimal distribution of material in a given design domain to achieve design objectives (e.g., high strength) while satisfying design constraints (e.g., low weight) [44]. With such capabilities, TO plays crucial role in many materials and structural design applications [4], [37]–[39]. However, many TO methods face the challenge of repeatedly computing an expensive objective function, which involves finite element (FE) solvers at each iteration [17]. These computational processes can be time-consuming and demand significant computational resources, especially for problems that involve nonlinear analysis, such as plasticity, viscoelasticity, and dynamics. To this end, ML models can act as surrogates to replace computationally expensive FE calculations in the objective function, thereby accelerating design optimization processes [45]–[47]. Moreover, ML-based models are computationally inexpensive and differentiable, making them suitable for gradient-based TO methods, which can further facilitate the optimization processes [48].

In addition to being used as surrogate models to replace expensive numerical models, ML can also be used to directly model the topology optimization process. In this approach, ML models take the design objective and constraints as input and give the final topology design as output. Various topology optimization problems, ranging from structures under static load [49] to conductive heat transfer [50], have been effectively tackled using CNN-based approaches. Deep generative models, such as GAN, can also be used to predict the optimal topology design, given design objectives and constraints. As reported in the literatures, GAN-based methods have demonstrated great successes in the optimization of material microstructures for desired properties [17], [51], [52]. Finally, reinforcement learning (RL) has also shown several successes for recurrent optimization problems, as documented in prior research [28], [53].

## 2.3  Design for manufacturing

ML-based methods are also advantageous in design for manufacturing (DFM) due to their strong modeling capability and computational efficiency. Assimilating process-structure-property (PSP) linkages is a crucial task in computational modeling for DFM but remains a challenging task. Advanced manufacturing techniques, such as metallic additive manufacturing, involve various physical and chemical processes that occur across multiple scales during the transformation of raw materials. As these processes are complex, traditional numerical approaches such as finite element analysis (FEA) or cellular automata can be expensive and inadequate for process optimization, monitoring, and control. In such cases, ML can be used to simplify the computational models by serving as surrogates or even replace the expensive numerical models entirely [54].

For instance, Tapia et al. [55] employed Gaussian Process (GP) to predict the depth of the melt pool in laser powder bed fusion of 316L stainless steel. The GP model was trained using data from high-fidelity physics-based numerical simulations and was able to accurately predict the melt pool depth of single-track prints based on inputs including laser power, velocity, and spot size. Furthermore, the use of GP allows for the assessment of uncertainty in predictions, which facilitates the making of robust and reliable design and control decisions [56]. In the same vein, Wang et al. [57] proposed a physics-informed data-driven modeling framework that can connect multiple data-driven surrogate models at different levels, allowing for the effective estimation of manufacturing uncertainty.

In brief, applying ML to DFM has great potential for improving the optimization and control of complex manufacturing processes. However, this research area is currently overlooked, and more effort is needed to advance the field further.

## 3    Challenges in machine learning for multiscale computational modeling and potential solutions

As we have seen so far, there have been reported in the literature notable successes of applying ML to multiscale computational modeling. However, there are still several critical obstacles that need to be surmounted before ML can be used in real-world applications. In this section, we discuss these challenges and offer potential solutions.

### 3.1    Data dependency

The use of ML for multiscale computational modeling presents several challenges that require attention. Although modern ML techniques like DL can achieve high predictive performance, they necessitate a considerable amount of training data. However, collecting such training data is often an expensive and tedious process, especially when dealing with experimental data. Synthetic data generated from numerical simulation may offer an alternative, but producing them can also be computationally demanding due to the complexity of multiscale problems [16], [58]. Insufficient training data can result in erroneous predictions or overfitting, reducing the generalizability and trustworthiness of ML algorithms.

Besides the amount of training data, the quality of data is also crucial for the success of ML algorithms. Training data with missing values or many outliers/noise can cause ML models to learn incorrect patterns, resulting in poor prediction performance, preventing the application of ML for real-world practice.

In the subsections below, we discuss several potential directions in this regard, including transfer learning and meta-learning.

#### 3.1.1    Transfer learning: utilizing prior knowledge from previous training

Training ML models for computational modeling requires a large amount of data, which can take an along time and effort to collect. Unfortunately, when a new problem arises, the same data collection process must be repeated, and the ML model needs to be retrained with the newly collected dataset. However, repeating the long and tedious data collection process is not desirable because collecting data for ML models used in multiscale computational modeling is often expensive. Moreover, retraining an ML model from scratch (Fig. 2(a)) may not always yield similar statistical performance compared to previously trained results. In such cases, one may consider utilizing prior knowledge gained from previous training sessions to reduce the need for obtaining a large training dataset for a new problem.

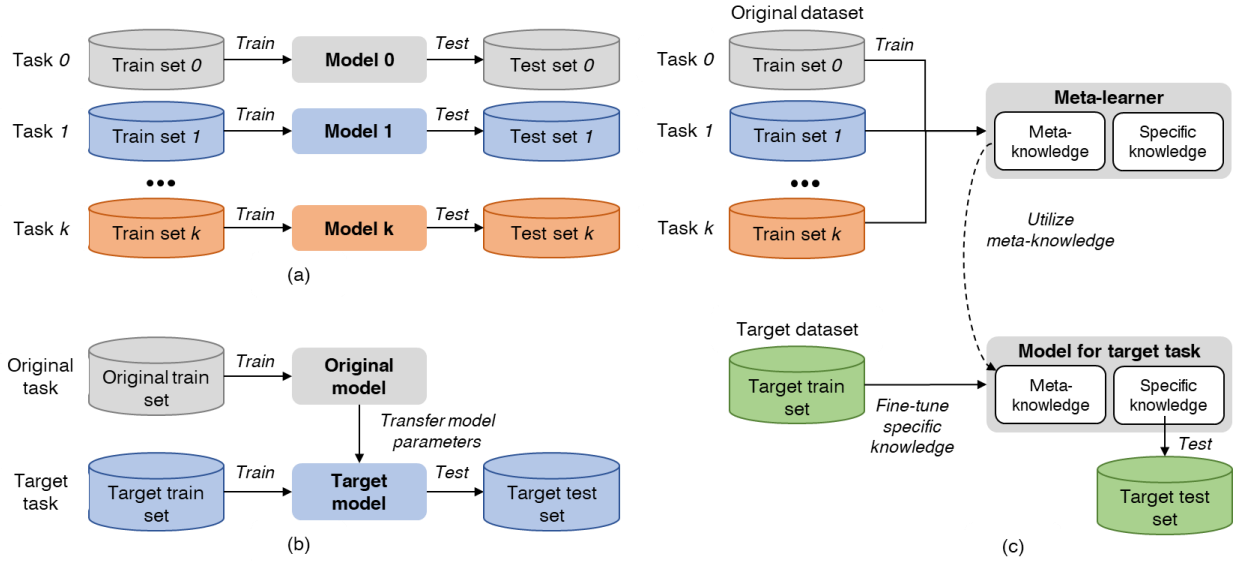

Figure 2. Three types of training ML models. (a) Training from the scratch for each task. (b) Transfer learning. (c) Meta-learning.

In this vein, transfer learning [59] (Fig. 2(b)) has proven to be successful in many generic ML applications, such as image recognition [60] and language modeling [61]. In transfer learning, model parameters of a pre-trained ML model, trained on a similar but different dataset, are used as the starting point for the current training session. In the new training session, relatively minor changes are applied to a subset of model parameters, while a big portion of model parameters can be "frozen" and not updated. This allows the ML model to fine-tune its model parameters to learn the new problem with only a small number of newly added data. Transfer learning can reduce the need for a large dataset while maintaining the prediction performance of the ML models. Previous studies have shown that transfer learning is effective in training ML models to predict the deformation of structures under static loads in new loading scenarios [62]. Transfer learning has also been found to be successful in predicting the response of energetic material microstructures when subjected to new shock loading conditions [34].

### 3.1.2 Meta-learning: learning generalized knowledge for quick adaptations

A step further from the idea of transfer learning is the more generalizable concept of 'meta-learning' (Fig. 2(c)). Meta-learning allows ML algorithms to quickly learn and adapt to new tasks by leveraging experience gained from solving similar tasks. In this context, transfer learning can be considered as a specific case of meta learning.

Conceptually, in meta-learning a meta-learner is constructed to learn generalized knowledge (meta-knowledge) across different learning tasks and utilize it as a prior to learn specific knowledge for each task, thereby maximizing the likelihood:

$$P(\theta_i|D_i, D_0) = \int_\phi P(\theta_i|\phi, D_i)P(\phi|D_0)d\phi \approx P(\theta_i|\phi^*, D_i)P(\phi^*|D_0) \qquad (1)$$

In Eq. (1) the meta-learner is represented by meta-learning parameters $\phi$ and the knowledge on a specific task $\sigma_i$ can be represented as a combination of the meta-knowledge $\phi$ and the specific knowledge $\theta_i|\phi$. The meta-knowledge is acquired via training on the reference dataset $D_0$ which maximizes the posteriori $\phi^* = \underset{\phi}{\mathrm{argmax}}\, P(\phi|D_0)$. In more detail, given the reference dataset $D_0$, the algorithm will first learn generalizable knowledge by leveraging the richness of the reference dataset $D_0$, and then use it as a basis to learn the specific knowledge $\theta_i^*$ for the specific task $\sigma_i$ using only a small data set $D_i$.

Previous works showed that meta-learning approaches could provide good prediction performance on unseen tasks despite being trained with relatively small datasets. For instance, Liu et al. [63] initialized physics-informed neural networks (PINN) with the Reptile algorithm, a meta-learning initialization method. The authors demonstrated that PINN can be trained more quickly with initialization guided by meta-learning algorithms than with the conventional randomized initialization methods. This new PINN training paradigm was successfully validated with a variety of forward and inverse test problems. In another study, Li et al. [64] proposed a physics-informed meta-learning (PIML) model for tool wear prediction problems in milling process modeling. The authors incorporated a physics-informed loss term that ensured the meta-learner learned the robust relationship between tool wear rate and cutting forces, thereby facilitating parameter estimation and enhancing the interpretability of the PIML model.

Previous examples have demonstrated that meta-learning is well-suited for the small-data context of ML in computational modeling of multiscale problems. Meta-learning allows for rapid adaptation to new operating conditions, such as new materials, geometric domains, or boundary conditions, without requiring a large training dataset. In addition, the "learning-to-learn" procedure can also aid to bridge the knowledge gap between simulations and real-world experiments. Currently, ML models are commonly trained with simulation data, as observational data from physical experiments is typically limited and expensive. However, there is a substantial difference between the idealized environment of numerical simulations and the real-world environment of physical experiments, which may cause ML models to perform poorly in real-world settings. In such situations, meta-learning can be used to rapidly adapt pre-trained ML models from numerical simulations to real-world environments. Specifically, a meta-learner can be trained to learn both meta-knowledge representing underlying physics laws and specific knowledge representing the operating environment of datasets, i.e., ideal assumptions of numerical simulation or real-world environments. At the deployment stage, meta-learning can then be utilized to quickly adapt ML models on ideal environment assumptions to real-world environments, using just a small amount of real-world experiment data. With this approach, ML models exclusively trained with numerical simulations can be quickly adapted to operate in real-world environments.

### 3.2 Discretization matters: the need for discretization-independent models

Applying ML for multiscale computational modeling also presents another challenge that is the dependence of ML algorithms on the discretization of the analyzed domain. Once trained, most of the common ML techniques, such as CNNs, are limited to a specific discretization setting (Fig. 3(a)) and do not generalize well for settings that differ from the discretization of the training data. This issue of discretization dependence poses a significant challenge for training ML models for multiscale computational modeling.

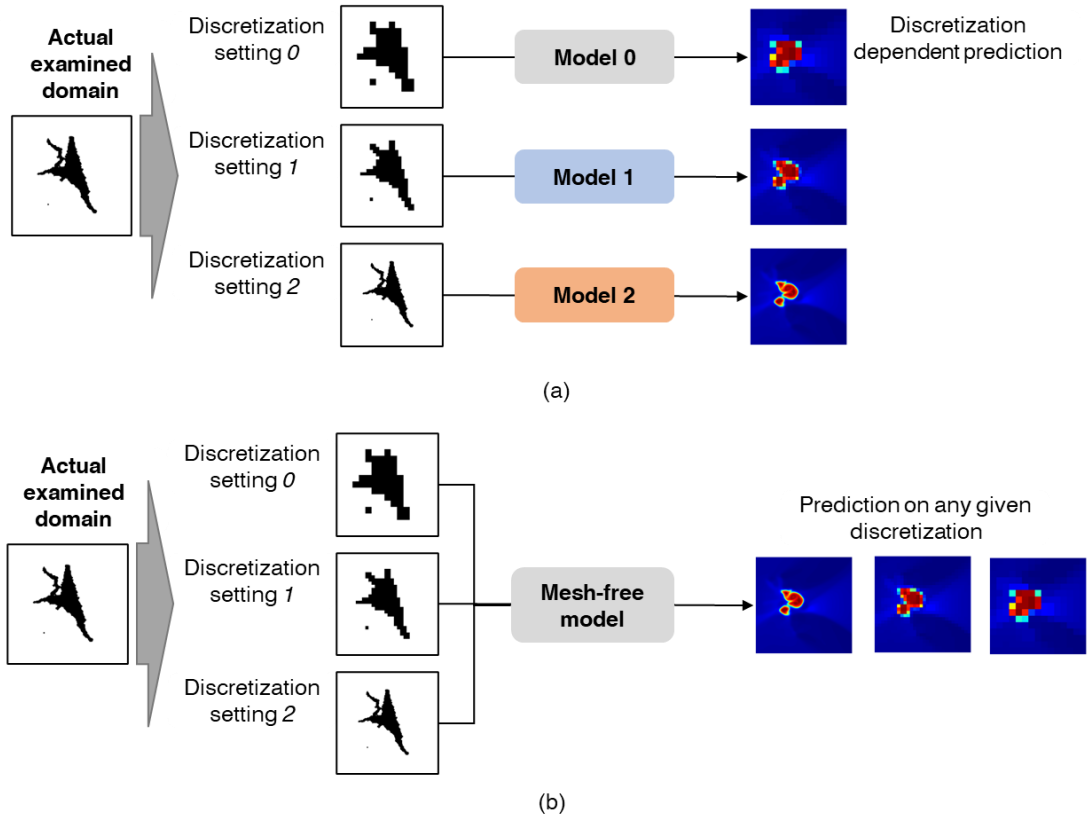

Figure 3: The difference between discretization variant (a) and invariant models (b). The discretization variant models are not generalized beyond the discretization setting of training data. Meanwhile, discretization invariant model can be trained on multiple dicretization settings and its prediction is not dependent on the discretization of input fields.

The dependency of ML algorithms on the discretization requires efforts to determine the optimal discretization setting that ensures convergence properties for all data samples in the dataset. However, the convergence property of a domain under examination heavily depends on its geometric characteristics, which notably vary across all data samples in the training dataset. Therefore, finding the optimal discretization setting is extremely costly and time-consuming. For this reason, developing discretization-independent (Fig. 3(b)) ML models is preferable as these models can make predictions regardless of the discretization of the examined domain.

Initial efforts have been devoted toward developing discretization-independent ML models for computational modeling of physical problems. For instance, Wandel et al. [65] transformed the discretized representation of state variable fields into continuous representations by interpolation with Hermite spline kernels. The Hermite spline constants were stored for each discrete grid point and could be handled by a CNN. This method successfully modeled several complex flow phenomena, such as Karman vortex streets, the Magnus effect, the Doppler effect, interference patterns, and wave reflections. Quantitative assessment also showed that using CNN on a continuous representation of state variables can improve the accuracy of ML models while still maintaining their computational efficiency advantage. In another work, Sun et al. [66] proposed the use of a continuous convolution kernel based on the Zernike polynomial for scalar field regression. This method exhibited an increase in prediction performance compared to the conventional approach with discrete graph convolutions, as well as the capability of making predictions on arbitrary geometric domains with different discretization settings. The utilization of neural operators is another approach that can feature discretization-independent characteristics [67]–[71]. In this approach, neural networks are trained to learn operators, namely *neural operators*, that map between infinite dimensional input and output function spaces. Since neural operators are trained to operate on function spaces, their predictions are not affected by discretization settings of examined domain while the accuracy of prediction is still assured [71].

### 3.3 Interpretability of machine learning models

ML has demonstrated great potential in multiscale computational modeling, as discussed in previous sections. However, there is still a significant concern about the trustworthiness of these predictions, which prevents their use in real-world applications. This concern arises from the "black-box" nature of ML predictions. ML models are often trained to optimize predefined objective functions that may not fully describe the real-world physics, leading to the possibility of ML models merely "memorizing" observed data but not being aware of or representing the underlying physical phenomena. As reported by Nguyen et al. [16], the physics-naïve architecture of ML models developed for generic purposes, despite having relatively high prediction performance, often fail to capture the underlying physical laws. Due to these reasons, it is crucial to ensure the trustworthiness and physics-awareness of ML predictions before utilizing these models for practical computational modeling.

The need to justify the trustworthiness of ML predictions necessitates the development of methods to interpret ML model predictions, i.e., understanding of the rationale behind ML predictions. Several initial efforts have been made to develop method for interpreting ML predictions in physical science. For instance, Nguyen et al. [16], used saliency map visualization to highlight "critical" microstructure locations that have a high impact on the creation of energy localization in energetic materials modeling. The interpretation results were compared to previous findings, and a high level of agreement between the saliency map identification and the literature on the characteristics of "critical" energetic materials microstructure was demonstrated. In another work, Lellep et al. [72] utilized Shapley Additive Explanation (SHAP) algorithms to investigate the main factors that affect the ML prediction in fluid dynamics. The authors demonstrated the applicability of the method in various fluid dynamic datasets, including the evaluation of plane Couette flow.

### 3.4 Centralized platform for data sharing and collaborations

Large databases, such as ImageNet [73], MNIST [74], and others, play a crucial role in the success of ML in computer vision and language modeling. Similarly, centralized databases offer a potential solution to the challenge of a shortage of high-quality training data for ML algorithms in computational modeling. Moreover, with the advancement of cloud computing technology, these platforms can be expanded for shared and collaborative learning, bringing together computational analysts, experimentalists, and ML researchers to facilitate effective collaboration. Unfortunately, such a platform is currently lacking.

Figure 4 depicts our proposed centralized platform for data sharing and collaboration in ML for computational modeling, which is inspired by OpenML [75], an example of a successful sharing platform in the ML community. The proposed platform consists of three main components. Similar to several available platforms for scientific data sharing [76], [77], our proposed platform includes an open scientific database for sharing and retrieving data. Additionally, ML model libraries are included, allowing for the sharing of previously successful ML models. Finally, the proposed platform includes an interactive environment that allows scientists to collaborate on a particular project. With the above features, the proposed centralized data sharing platform can create a common format and protocol to facilitate data and model sharing. Because engineering data are generally heterogeneous and can come from multiple sources, having such a sharing platform will enrich the dataset that can be used to train ML models, thus helping to overcome the shortage of training data.

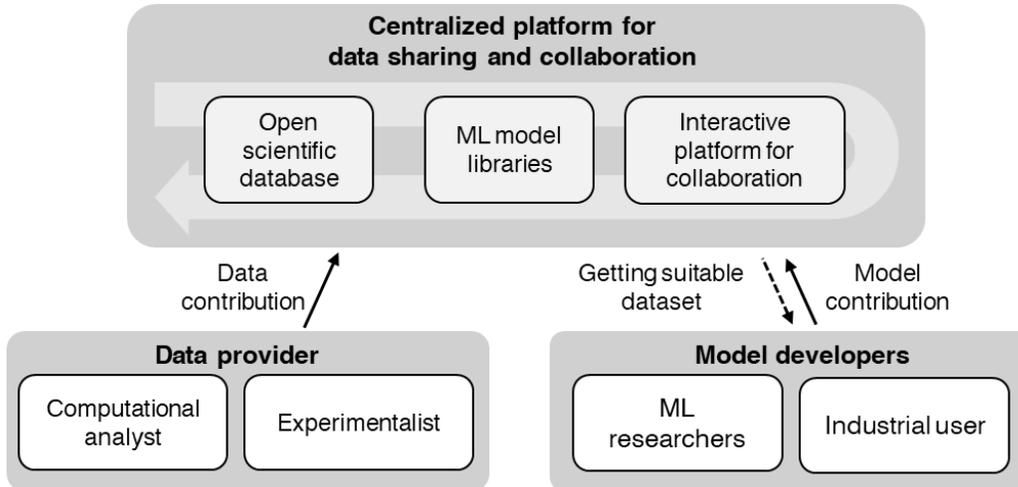

Figure 4: A centralized platform is necessary for data and model sharing as well as supporting collaboration between data providers and model developers.

Another important feature of the proposed sharing platform is its ability to facilitate collaborations via an interactive platform. In this collaborative platform, several most suitable datasets and ML models can be suggested to researchers for specific problems. Researchers can then decide whether to utilize the suggested datasets and models or work with their own. Additionally, this platform allows ML developers to request or recommend additional data samples for simulation or experimentation based on the performance of the model on the current dataset. Consequently, experimentalists or computational analysts can confirm the validity or suggest alternate data samples based on their expertise.

Compared to the conventional method of data sharing, this collaborative mechanism offers several advantages. First, this mechanism can reduce the time to collect and preprocess training data since the quality of training datasets suggested by the collaborative platform has been justified via previous training sessions. Additionally, through this mechanism, ML developers can gain access to meaningful and valuable data, thereby improving the resulting ML model's predictive performance. Finally, this mechanism aids to justify the robustness of ML algorithms. In computational modeling, ML models are typically trained on a single dataset, which can cause ML predictions to be biased toward the training dataset. With this collaborative environment, an ML model can be trained and validated on multiple different datasets, enhancing its robustness and generalizability.

## 4  Conclusion

Multiscale computational modeling plays a critical role in many mechanical engineering applications, such as computational mechanics, mechanical design and optimization, and manufacturing/process modeling. Despite the promise of capturing complex physics phenomena, the progress in multiscale computational modeling research is hampered by the high computation cost. To this end, ML can be a viable option that can aid in the analysis of multiscale physical systems by either augmenting or serving as a surrogate for multiscale computational models.

However, several obstacles must be surmounted to successfully apply ML to the computational modeling of multiscale physical systems. The first pressing issue is big data dependency, which requires immediate attention. Unlike traditional ML problems, such as image or text classifications, collecting scientific data is difficult and expensive, particularly for multiscale physical systems. The lack of large, high-quality training datasets can significantly reduce the generalizability and robustness of ML models. Investigating the use of prior knowledge through physics-informed ML, transfer learning, and meta-learning are all viable options to overcome this challenge. Another issue is dealing with the dependency of ML algorithms on the discretization setting of the examined domain. To overcome this issue, developing ML on continuous representation or applying neural operators are potential future directions. In addition to improving the statistical performance of ML-based computational models, it is equally important to validate the physics-awareness and trustworthiness of ML predictions. Finally, the development of a centralized platform for data and model sharing is necessary to facilitate the development of algorithms and foster collaboration in the field of multiscale computational modeling.

Aside from a few pioneering works, the above-mentioned issues are still understudied. Therefore, there is an opportunity for future research to explore better solutions to these challenges. Once these obstacles are overcome, ML has the potential to significantly accelerate the development of multiscale computational modeling and facilitate the discovery of novel and more effective mechanical systems.

**Acknowledgement**

This paper is based upon work supported by the U.S. Air Force Office of Scientific Research (AFOSR) Multidisciplinary University Research Initiative (MURI) program under Grant No. FA9550-19-1-0318 and by the National Science Foundation (NSF) Designing Materials to Revolutionize and Engineer our Future (DMREF) program under Grant No. 2203580.